# Nahid: AI-based Algorithm for operating fully-automatic surgery

Sina Saadati[1]

## Abstract

In this paper, for the first time, a method is presented that can provide a fully automated surgery based on software and computer vision techniques. Then, the advantages and challenges of computerization of medical surgery are examined. Finally, the surgery related to isolated ovarian endometriosis disease has been examined, and based on the presented method, a more detailed algorithm is presented that is capable of automatically diagnosing and treating this disease during surgery as proof of our proposed method where a U-net is trained to detect the endometriosis during surgery.

**Keywords:** Deep Learning, AI in Medicine, Image Processing, Computerizing Surgery, AI.

## 1 Introduction

Nowadays, computer technology and artificial intelligence have revolutionized all aspects of the science. The computerization of scientific, industrial, clinical, and commercial processes leads to speeding up the process, increasing accuracy, and reducing costs. In the field of medicine, artificial intelligence methods have been able to perform on par with or even better than humans in many clinical purposes. The introduction of artificial intelligence into surgery can increase the speed of surgery and thus reduce the amount of bleeding and the risk of infection. This goal can be achieved in two ways, fully automatic or computer-assisted way.

Computer methods based on software, high computing power, and artificial intelligence have significantly improved human ability. Today, computers can process very large amounts of data and information at a very high speed and with sufficient accuracy. This feature includes image processing and machine vision algorithms. In such a way artificial neural networks (ANN) that try to simulate the brain of humans or animals can see the images and conceptualize them. Hence, this category of calculations is also called deep learning. The introduction of technology and specifically, computers to medicine and surgery makes it possible for doctors and researchers to carry out scientific, research or clinical processes with greater speed and less cost.[8,9,10,12,13]

In this article, focusing on the surgery for isolated ovarian endometriosis, a method is provided that enables fully automatic surgery by computer and artificial intelligence. Endometriosis is a common disease among women that can lead to pain in the abdominal and pelvic areas and increase the risk of infertility. In this disorder, tissue called endometrioma grows abnormally around the uterus. This phenomenon can be seen in different areas such as ovaries, uterus, intestines, etc. Isolated ovarian endometriosis is a type of disease in which endometrioma tissue is observed only in the ovary. This type of disorder is statistically less common than other conditions in which endometrioma tissue is found in several places. This disorder can be detected using methods such as the use of ultrasound waves or magnetic resonance imaging (MRI) that are performed before surgery. Also, the diagnosis and treatment of this disease can be done by minimally invasive surgery or laparoscopy. In some cases, endometrioma tissue is removed by thermal waves or laser during surgery. Recent research has shown that using a laser for removing endometriosis has fewer risks compared to other methods. [11]

Laparoscopy or minimally invasive surgery is a modern method in which only limited areas of the body are split and through which tools are inserted into the body to watch, hold, observe, or cut inner organs. This surgical method has an accuracy equal to or greater than the traditional surgical method, and since the connection between inside and outside spaces of the body is minimized, it leads to a reduction in the risk of infection. It has also advantages like less bleeding and pain. Today, this method has been combined with robotic technology and specialists are able to operate on patients from a long geographical distance. This article will take another step in the field of modern surgery. In the second chapter, the works related to this research are briefly introduced. Then, in the third chapter, a comprehensive method for fully automatic surgery will be presented. Software architecture and requirements engineering are described in this section. In the fourth chapter, the opportunities and challenges of this method will be discussed. In the fifth chapter, isolated ovarian endometriosis is focused, and based on the proposed method, a more detailed algorithm will be described to perform the surgery in a fully automatic way. Since the algorithm presented in this section is based on machine learning and computer vision methods, a neural network model is introduced that is capable of diagnosing and treating isolated ovarian endometriosis. Finally, in the sixth chapter, the conclusion will be discussed.

[1] Department of Computer Engineering, Amirkabir University of Technology, Tehran, Iran, Sina.Saadati@aut.ac.ir

## 2 Preliminaries

In recent years, modern technologies have enabled significant growth in the fields of biology, medicine, and surgery. Nezhat et al. have examined the recent advances in engineering sciences, especially artificial intelligence and energy production. Then, they predicted that in 2050, a large fraction of surgical procedures will be performed by computers. In this research, obstacles such as the possibility of bias of patients and even the community of doctors and surgeons with the introduction of technology into medicine have been pointed out.[5] In a research focusing on minimally invasive surgery and robotic surgery, Mascagni et al. pointed out the importance of machine vision issues and the availability of image data to advance medical research. He has classified the methods of computerization of medical services; However, surgery is not mentioned in it as a completely automatic method.[3] Maddzadeh et al. have presented a dataset that can be used in training machine learning models and especially artificial neural networks. The purpose of this data collection is only the production of intelligent assistants in medicine and surgery.[2] In a research, Scheikl et al. investigated different architectures for the semantic segmentation of images related to endometriosis. In this research, in order to meet the minimum speed of calculations, the processing of each image in a period of less than forty milliseconds is recommended. This characteristic leads to the image processing in real-time mode with the use of the surgeon's assistant.[6] Visalaxi et al. reported a research, using ResNet50 architecture to produce an artificial intelligence model that can detect the presence or absence of endometriosis disorder in images; This model only responds in a binary way and does not provide the possibility of localizing the endometrioma tissue. [7] One of the challenges of this research is that endometriosis disorder is considered an unclassifiable disease. This disease has different types that are treated in different ways. Referring to this point, in addition to facilitating the process of training and evaluating artificial intelligence, it can lead to an increase in the accuracy of the results. In a research, Hong et al. have labeled several sets of data related to laparoscopic images. In this research, the classification and categorization of data has received less attention. [1] In a study, Naqvi et al. suggested that researchers use the Glenda dataset, which includes laparoscopic images related to endometriosis. [4] In addition to the fact that this dataset presents a very small and limited number of images (less than 400 samples), no difference has been made between the location of endometriosis. So far, it can be observed that in only twenty-six samples there is a possibility of watching ovary and in this subgroup, only sixteen cases include endometrioma located on the ovary. In the present study, in addition to Glenda's data, other data including ovarian endometriosis have been used.

## 3 Nahid: Artificial Intelligence capable of managing surgery

In this section, an AI-based algorithm named Nahid is presented for the complete automatic operation and management of surgery. This algorithm is implemented based on Sina algorithms and Sina tree data structure. To provide a suitable and efficient algorithm that enables a computer to operate a surgery, we should first consider how humans are capable of doing surgery. What leads a person to perform surgery is related to his knowledge and vision. In this claim, knowledge means the human ability to diagnose and distinguish organs and disorders. Vision is also an essential factor in the surgery. Using the combination of vision and knowledge, experts are able to identify the abnormalities in the body and remove them. As a result, to produce artificial intelligence that can appear as a surgeon or surgeon assistant, it is necessary to produce these two features in the form of software. In this section, different methods will be presented to explain how to inject the knowledge into the computer.

The first method, which seems perfect and challenging to develop, is to produce software for modeling and simulating the human body. It means software that simulates all the patient's organs by receiving the characteristics of the patient, such as age, gender, weight, disorders, etc. The tool should be able to provide mobility and movement of organs virtually to the user. Also, the developed software should be able to perform routing safely and optimally. For example, calculate the shortest path to reach from the belly button to the uterus in such a way that the least damage is done to the internal organs. In this method, features such as the color of the internal organs, the elasticity of the organs, and physical actions like resistance to cutting and heat should be considered. The difficulty of physical equations related to elastic organs and their computational complexity makes this method challenging and hard to develop. Another method is to probabilistically define points of the body as a three-dimensional graph; So that each point expresses the probability of presence of a certain part of the body. For example, in the coordinates like (x,y,z) the heart or uterus can be found with high probability. This is useful where the camera (or robotic arms) are located inside the body and computer needs to find out what is probable to be seen(or grasped). In any case, we call the software that enables the positioning of the organs and their related disorders by considering the conditions "human geometric model". In addition to the human geometric model, it is necessary for the computer to be aware of the necessary steps to manage surgery. This can be defined by an algorithm and implemented by a computer. These capabilities make it possible to inject the knowledge needed for surgery into the computer.

In addition to knowledge, it is necessary to develop the necessary AI models to conceptualize and recognize different organs in different conditions. In this research, unlike other deep learning models, it is suggested that researchers separate the artificial intelligence models based on the surgical states. Glenda's data collection includes a small collection of images that are prepared during minimally invasive surgery for endometriosis patients. Endometriosis is a disorder that can occur in different areas that have a different appearance and geometry. Figure 1 separates these areas. By classifying the disorder of endometriosis based on its appearance location, we will be able to collect images that are less different in terms of geometry and appearance. In this way, this opportunity is provided to train a model with much higher accuracy even with limited data sets. With this approach, the time and energy required for training and test processes, as well as the application of artificial intelligence models for image processing, will be optimized. The separation of data and artificial intelligence models based on categories of disorders not only leads to the optimization of technical criteria but also increases the transparency of the behavior of neural network models. This point is important from the point of view that in medical applications, having black box software components due to the importance of the treatment processes can lead to questioning the whole treatment method. In this research, we call this suggestion the "principle of situation separation".

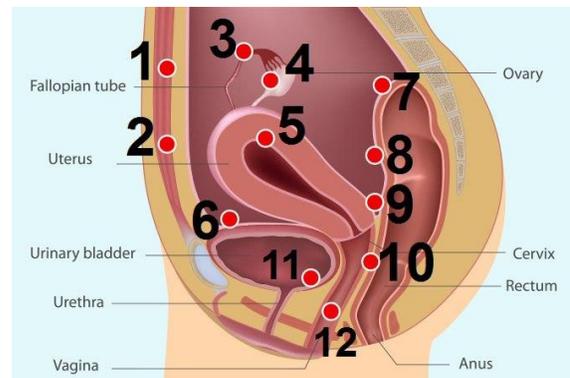

Figure 1, Endometriosis probable locations.

Computer vision refers to the use of artificial neural network models that are capable of performing processing to understand and separate image content based on boundaries, meaning, etc. For this purpose, we need a dataset including normal images and labeled masks. For example, in Figure 2, a pair of images can be seen including a real laparoscopic image and its associated labels. [1] This image, provided by Hong et al., provides the ability to train a neural network and use it to understand and segment the newer images. In this research, based on the principle of situation separation, it is recommended that each stage of surgery should have an independent model that is trained based on separate datasets that correspond to the same state and location of surgery.

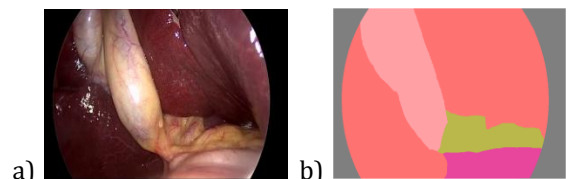

a)  b)

Figure 2, An image segmentation example[1]

Observing the situation separation principle has another advantage, which is related to the risk of data overfitting. In their research, Nazhat et al. pointed out the difference between treatment and surgery methods between two patients suffering from one disease with similar features and, as a result, recommended standardization.[5] In this research, due to the fact that robotic surgery has provided the ability to manage surgery for a wide range of disorders, we suggest that certain and safe paths for the movement of the camera or robotic arms should be considered based on the human geometric model. Considering this standardization, we will be able to observe the inner organs from the same angle at each state of the surgery. It is important to consider other characteristics such as age, weight, place of disorder, etc. One of the advantages of following this principle is the minimization of the difference between training and test data. With these conditions, overfitting in the process of machine learning will not be dangerous.

In the following, with the aim of combining the human geometric model with machine vision, a computational architecture is presented that can manage surgery. First, an algorithm is presented that uses several machine vision models to detect and understand images. Then, the surgical management will be explained.

| Sina Algorithm: |
|---|
| Input: an imae of minimally invasive surgery (1) along with the operation situation that includes the exact location of the camera in the body (2). Output: detection of all the elements in the image. |
| 1- Based on the computer vision model, which is related to the surgery situation, proceed to the semantic segmentation of the image and save the result as SegmentedImg. 2- Using a suitable algorithm, the borders in the image are detected and saved in the form of an image called EdgeDetectedImg. 3- Due to the high accuracy of EdgeDetectedImg compared to the borders found in SegmentedImg, the borders in EdgeDetectedImg should be considered as real borders. 4- For each area detected in EdgeDetectedImg, the pixels in SegmentedImg should be separated. Then, based on the highest number of pixels in each area, type of the areas must be labeled. 5- Save the result as the output of the algorithm. |

In order to increase the processing speed, the Sina algorithm can be executed in a parallel way. Figures 3, a, and 3, b show the second step of Sina's algorithm. As you can see, it is possible to demarcate organs using edge detection. However, edge detection algorithms are only able to recognize the edges in an image. To recognize and understand the type of each organ, it is necessary to use the appropriate neural network model. For this purpose, in the third step of Sina's algorithm, based on the physical position of the camera(situation of the surgery), the appropriate machine vision model is selected and labeling is done according. In this algorithm, edge detection is used to remove the predictable noises. Finally, the computer can understand and separate the organs and their related disorders well at every stage of the surgery. The proposed software architecture for implementing the Sina algorithm is shown in Figure 4.

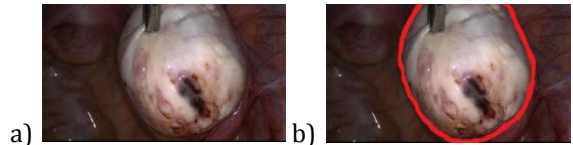

Figure 3, Edge Detection for finding borders.

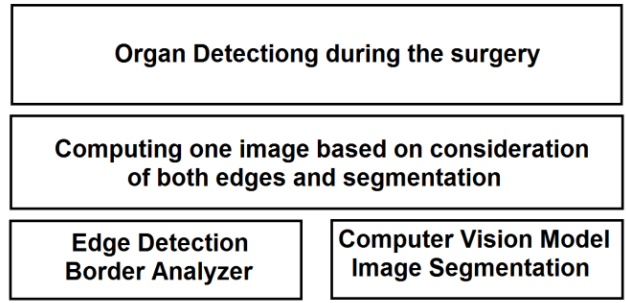

Figure 4, Proposed architecture for executing Sina Algorithm

Because in each stage of surgery according to the principle of state separation, it is essential to use an independent computer vision model, it is vital to store these models in a suitable data structure. In this research, a tree data structure is presented which tries to implement the human geometric model and respect the separation principle at the same time. We call this data structure the Sina tree. Each node of this building has a three-dimensional coordinate that is related to the position and angle of the camera in the body. In other words, all the images obtained from minimally invasive surgery can be divided based on the location and angle of the camera, and then the training and production of independent computer vision models can be done. Finally, each model is stored in a corresponding node in the graph. Also, the edges between the nodes are considered the path of the camera movement. By implementing the Sina tree, we will have much simpler, clearer, more accurate, and optimal models, and at the same time, it is possible to move from one position (or surgery situation) to another position of the body (or surgery situation). It can be easily done by following the graph edges. The existence of the Sina tree can also require compliance with the applied standards. Figure 5 shows an example of a Sina tree. Based on this data structure, the camera is inserted into the abdomen from the navel area and can reach the uterus or ovaries by navigating a direct path. It is possible to increase the accuracy and security of the operation by creating intermediate nodes on each edge. In this image, nodes are marked with red color, and edges are shown as green lines.

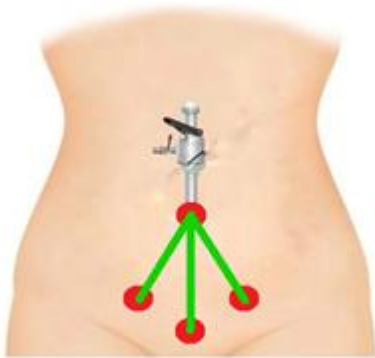

Figure 5, An instance of Sina Tree data structure

## 4 Challenges and Advantages

In this section, the advantages and challenges of surgery with the help of artificial intelligence from a social, research, education, and clinical services perspective will be examined. Fortunately, the rapid growth of computer technology and artificial intelligence has been widely welcomed by society. However, based on the sensitivities in treatment issues, a wide range of patients may resist accepting treatment with the help of artificial intelligence. To overcome this challenge, it is recommended that artificial intelligence applications have an expert supervising the treatment and surgery process. This approach can also resolve legal issues related to treatment. It is suggested that international institutions should also take necessary measures to promote public culture to trust technology even in medical matters so that the untrue fear of this technology can be removed. Computer error is much lower than humans, and as a result, trusting artificial intelligence seems logical. However, in this context, the correctness of artificial intelligence software and models must be evaluated and proven. This issue requires more research in this field. The results of this research will be effective not only in medical services and clinics but also in education. As claimed by Mascagni et al., many images and datasets related to the training of neural network models can be used for training by institutions and universities. Also, models and algorithms related to automatic surgery can be used to produce suitable simulators for training or research. Finally, by using the Nahid algorithm, we can increase the speed of surgery and reduce the error. With this measure, treatment costs will be also minimized.

One of the important challenges can be seen in the training of neural networks. In this regard, there is a strong need for data and images related to minimally invasive surgery. Unfortunately, the presented datasets do not provide enough information and have not followed the principle of situation separation. In this context, it should be noted that the imaging settings and configuration during minimally invasive surgery must be accurately recorded. For example, the intensity of light or the type of light used during imaging must be recorded.

## 5 Isolated ovarian endometriosis surgery using the presented algorithm

In this section, a special form of Nahid's algorithm for surgery of isolated ovarian endometriosis disorder is presented. This algorithm is described using the flowchart in Figure 6. In this process, the way to use the entities related to Nahid algorithm, such as Sina algorithm or Sina tree, are depicted with appropriate details. Also, the Sina tree for this type of surgery is drawn in Figure 7. In this diagram, each state is comprehensively expressed by a three-row rectangular. The first row of each column indicates the state or stage of the surgery, the second row includes the location of the camera, and the third row indicates the task that artificial intelligence must perform at that state. This chart will contain much more detail in a real problem. For example, it is necessary to save the position of the robot's arms in addition to the camera position.

As can be seen, in this process, three nodes are considered to move from the navel area to the patient's left ovary. Each node includes a powerful computer vision model that is able to specifically detect all organs or disorders in that state. Based on the knowledge gained from the computer vision model and the Sina algorithm, as well as the Sina tree data structure, it is possible to move forward or backward in the body during the operation easily. This algorithm can be designed in other ways for other kinds of diseases, But the principles of its definition and design are the same as what was explained in this article.

One of the most important steps in this process is related to the diagnosis of endometrioma tissue located in the ovary. The recognition of this tissue by computer vision makes the robot behave correctly in targeting the radiation of thermal waves or lasers. In this research, to prove the feasibility and correctness of implementing Nahid's algorithm, it has been done to produce and evaluate a neural network model to diagnose endometrioma on the ovary. In this context, a U-NET neural network architecture with a sigmoid activation function has been used. The input of this network is defined as a 128 x 128 matrix that represents the images of the ovary. Its purpose is to provide an image focused on the ovary and to calculate the exact location of the endometrioma disorder by artificial intelligence. In this context, due to the insufficient data of the Glenda dataset, new data has been collected and added to the dataset. Also, by using the image rotation technique, the number of data elements has increased to four times. It should be noted that due to the existence of standards in surgery, there is no possibility of the camera being uneven during surgery. In this research, this method was used only to overcome the problem of data limitations. Not using this technique and, on the contrary, increasing the number of data will lead to an increase in the accuracy of the model. Finally, a model with reasonable accuracy was produced. Table 1 reports the output of evaluation criteria on this model. The results of the evaluation of this model state that if the number of data increases and the principles defined in this article are followed, it is possible to fully automate and computerize surgery.

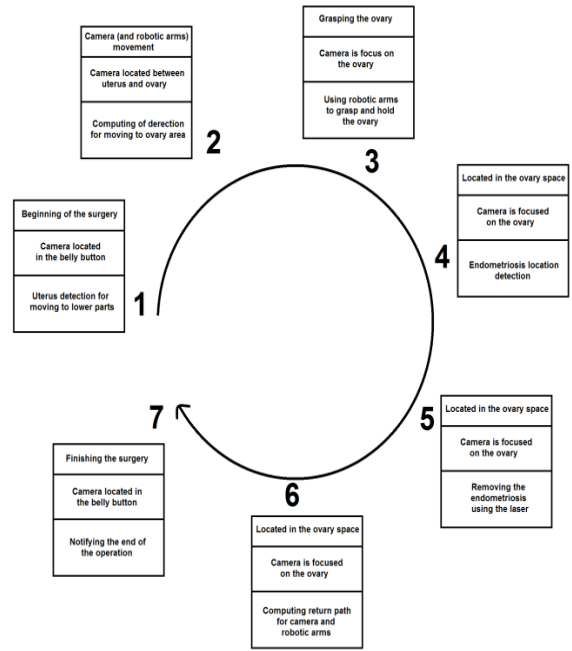

Figure 6, Flowchart of the proposed method of surgery

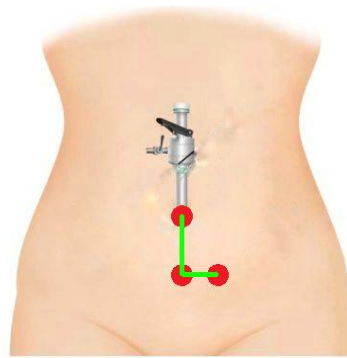

Figure 7, An example of Sina Tree for isolated ovarian endometriosis.

Table 1, Evaluation of ovarian endometriosis location detection model.

| Metric | Train Loss | Train IoU score | Val Loss | Val IoU score |
|---|---|---|---|---|
| Value | 2.85e-04 | 0.9995 | 0.0122 | 0.9759 |

Figure 8 shows some examples of images used as datasets. In these images, the ovary organ is considered focused. It is possible that by other neural network models, in the states before reaching the ovary, the ovary was separated from other organs, and based on the result, endometrioma tissue was diagnosed at the current state. Also, it is possible to prepare images from different angles of the ovary based on the human geometric model and use them for artificial intelligence processing to improve the accuracy of calculations. In the following, the calculated output of artificial intelligence (a) and the expected output (b) for training data (Figure 9) and test data (Figure 10) have been compared.

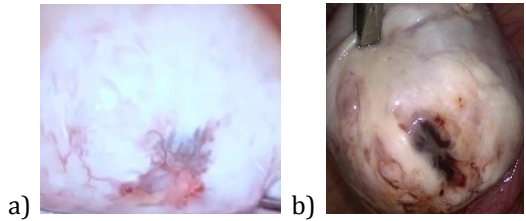

Figure 8, Two instance of data used to train the endometriosis location detector model.

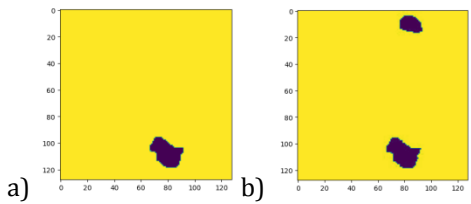

Figure 9, Comparison of predicted and true mask of the model.

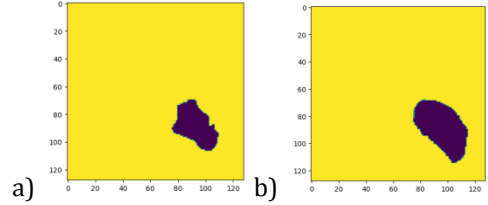

Figure 10, Comparison of predicted and true mask of the model.

# 6 Conclusion

In this research, for the first time, how to perform surgery by artificial intelligence was described. By describing the Nahid algorithm, which consists of the Sina algorithm and the Sina tree data structure, an attempt was made to provide a comprehensive, efficient, and accurate method for surgical management and operation by artificial intelligence based on a geometric model of humans and observing the principles of situation separation. Then, focusing on surgery related to isolated ovarian endometriosis disorder, how to use the Nahid algorithm for this type of surgery was explained. Also, by producing and evaluating a model based on an artificial neural network that detects and localizes endometrioma tissue in the ovary, it was proved that by following the Nahid algorithm, it is possible to produce artificial intelligence to navigate the operation and this process is reliable. The obtained accuracy, the lack of vulnerability of the model against the phenomenon of overfitting based on the principle of separation of the situation, and attention to the limitation of the datasets in this model prove the reliability of the proposed method.

# 7 Declarations

## 7.1 Appreciation

## 7.2 Funding

### 7.3 Conflicts of interest/Competing interests:

Not applicable.

### 7.4 Availability of data and material:

Not applicable.

### 7.5 Code Availability

You can access our library in Github: https://github.com/sinasaadati95/Model